\documentclass[lettersize,journal]{IEEEtran}
\usepackage[utf8]{inputenc}
\usepackage[T1]{fontenc}
\usepackage{kotex} 

\usepackage{graphicx}
\usepackage{multirow} 
\usepackage{xcolor}
\usepackage{amsmath,amssymb, bm}
\usepackage{booktabs}
\usepackage{cite}
\usepackage{subfigure}

\usepackage{array}

\newcommand{\ours}{\textit{CURA}}

\title{CURA: Size Isn’t All You Need – A Compact Universal 
Architecture for On-Device Intelligence}

\author{
  Jae-Bum~Seo$^{*}$,
  Muhammad~Salman$^{*}$, and
  Lismer~Andres~Caceres-Najarro
  \thanks{*Jae-Bum Seo and Muhammad Salman contributed equally to this work.}
  \thanks{This work was supported in part
    by the National Research Foundation of Korea under Grant NRF-2021R1I1A1A01041257 and in
    part by the research fund from Chosun University, 2024.
    (Corresponding author: Lismer Andres Caceres-Najarro).}
  \thanks{J. B. Seo and L. A. Caceres-Najarro are with the Department of Computer Science, Chosun University,
       Gwangju 61452, Republic of Korea. Email: sjb258@chosun.ac.kr; andrescn@chosun.ac.kr.}
  \thanks{M. Salman is with the Faculty of Computer Science and Engineering,  Ghulam Ishaq Khan Institute of Engineering Sciences and Technology, Topi, Sawabi, Pakistan. Email: salman@nsl.inha.ac.kr}
}

\begin{document}

\maketitle

\begin{abstract}
Existing on-device AI architectures for resource-constrained environments face two critical limitations: they lack compactness, with parameter requirements scaling proportionally to task complexity, and they exhibit poor generalizability, performing effectively only on specific application domains (e.g., models designed for regression tasks cannot adapt to natural language processing (NLP) applications). In this paper, we propose \ours{}, an architecture inspired by 
analog audio signal processing circuit
that provides a compact and lightweight solution for diverse machine learning tasks across multiple domains.
Our architecture offers three key advantages over existing approaches: (1) Compactness: it requires significantly fewer parameters regardless of task complexity; (2) Generalizability: it adapts seamlessly across regression, classification, complex NLP and computer vision tasks; and (3) Complex pattern recognition: it can capture intricate data patterns while maintaining extremely low model complexity.
We evaluated \ours{} across diverse datasets and domains. For compactness, it achieved equivalent accuracy using up to 
2,500 times 
fewer parameters compared to baseline models. For generalizability, it demonstrated consistent performance across four NLP benchmarks and one computer vision dataset, nearly matching specialized existing models  (achieving F1-scores up to 90\%). 
Lastly, it delivers superior forecasting accuracy for complex patterns, achieving 
1.6 times
lower mean absolute error and 
2.1 times
lower mean squared error than competing models.
\end{abstract}

\begin{IEEEkeywords}
Lightweight model, low parameter count, MLP-based architecture, on-device artificial intelligence, residual gated MLP. 
\end{IEEEkeywords}

\section{Introduction}
Modern consumer electronics such as smartphones, laptops, and 
internet of things (IoT) 
devices now feature advanced hardware components, including memory storage, 
central processing units (CPUs),
and special 
artificial intelligence (AI)
chips called 
neural processing units (NPUs)
\cite{silvano2025survey}. 
This advanced hardware architecture enables these devices to accommodate on-device 
AI,
which brings computational intelligence directly to end-user devices rather than relying on remote cloud infrastructure \cite{liang2020ai}.
Integrating AI capabilities at the device level yields numerous significant advantages, such as improved user experience through reduced latency  \cite{thota2024optimizing},
enhanced privacy protection through local data processing \cite{dhar2021survey},
and improved operational reliability by minimizing dependence on network connectivity\cite{tekin2024review}.
These advantages have driven the widespread adoption of on-device AI across diverse platforms, including smartphones (e.g., Apple’s Face ID, Google Pixel’s real-time speech recognition), wearables (e.g., Apple watch health monitoring, Fitbit SmartTrack), vehicles (e.g., Tesla’s full self driving), smart home devices (e.g., Amazon Echo), and industrial systems (e.g., Siemens’ SIMATIC edge for predictive maintenance) each harnessing on-device AI to deliver domain-specific benefits. 
With ongoing progress, on-device AI's will become a core component of next-generation intelligent systems across numerous sectors \cite{shi2020communication}.

In practice, most existing devices possess limited computational resources for real-time data processing. This limitation necessitates the development of lightweight architectures through model optimization strategies such as model compression, efficient architectural design, and hardware acceleration. Although notable advances have been made such as MobileNet~\cite{sandler2018mobilenetv2} for computer vision and TinyBERT~\cite{jiao-etal-2020-tinybert} for natural language processing, these lightweight models have several limitations. First, they often require extensive pretraining on large datasets, which is computationally expensive. Second, they typically depend on task-specific fine-tuning, which reduces their flexibility across applications. Third, they involve non-trivial architectural modifications, which makes them harder to adapt to highly constrained environments.
More recently, transformer-based models have shown 
state-of-the-art (SOTA)
performance in various domains, including vision and language. However, they consume significant memory and computational resources~\cite{NIPS2017_3f5ee243, NEURIPS2021_4cc05b35}, making them ill-suited for deployment on a device. Even alternatives that eliminate self-attention such as 
gated multilayer perceptron (gMLP)
~\cite{NEURIPS2021_4cc05b35} and 
residual MLP (ResMLP)
\cite{touvron2022resmlp} still require millions of parameters to match the performance of heavier models. 
A recent lightweight architecture, the time series mixer
(TSMixer), adapts existing patch-based and channel-mixing methods for time series prediction \cite{chen2023tsmixer, ekambaram2023tsmixer}; 
however, it lacks generalizability across diverse tasks and its parameter count scales with task complexity, which limits its efficiency in broader applications.



This paper proposes an ultra-lightweight and highly generalizable neural network backbone architecture called 
compact universal residual architecture (\ours{}),
specifically designed for on-device AI environments. The architecture of \ours{} is inspired by the theory of analog audio signal processing circuits, which is basically comprised of five main components. First, the 
voltage-controlled amplifier (VCA), which modulates the input signal strength based on importance. Second, the 
analog mixer (e.g., an op-amp summer), which works like a blending control (e.g., a dry/wet knob in audio equipment), where it combines the original (raw) signal with the processed one, so that important details from both are kept.
Third, the
nonlinear amplifier (e.g., an op-amp with diodes in the feedback loop), allows strong signals to pass while clipping extreme values and suppressing weak ones\footnote{Instead of treating all parts of the signal equally, it behaves differently depending on the signal’s intensity: Very weak signals are often ignored or suppressed (to reduce noise), Medium-strength signals pass through more naturally, and very strong signals are clipped or limited, preventing them from becoming too dominant or distorted}.
Fourth, the 
bandpass filter, passes only a desired range of frequencies and rejecting the rest (akin to extracting localized patterns). Finally, the 
buffer stage (e.g., BJT emitter follower) adjusts the signal level and impedance for clean delivery to the output stage.
\ours{} adopts the above signal processing--inspired architecture that achieves the intended computational functionality with minimal complexity. 
The 
gating unit, 
functioning as a 
VCA, 
dynamically suppresses unimportant neurons during training by modulating signal strength based on contextual importance. This targeted amplification drastically reduces parameter requirements and eliminates the need for maintaining redundant pathways. 
The residual combinational unit, 
akin to an analog mixer, preserves essential signal flow and prevents gradient degradation without additional overhead. The 
non-linear activation unit
acts as a non-linear amplifier, controls neuron output even when gates are nearly closed\footnote{For example, when the gate value is 0.1 and the residual value is 0.05, simply multiplying them gives 0.1 × 0.05 = 0.005—a tiny number that essentially shuts down the neuron. Our proposed method fixes this by adding back the original residual value (0.05), preventing the signal from becoming too weak and keeping information flowing through the network.}. 
%
The nonlinear activation unit introduces differentiable nonlinearity, akin to a nonlinear amplifier, enabling adaptive feature transformation and enhancing representational capacity.
The filter unit
acts as a bandpass filter, to capture key patterns efficiently with far fewer parameters than multi-head attention or 2D convolutions. Lastly, 
the output projection unit
behaves like a buffer stage, 
forwarding the signal to the next layer without introducing corrective complexity. 

To the best of our knowledge, no existing AI architecture for resource-constrained devices is simultaneously compact (requiring significantly fewer parameters), generalizable across diverse tasks (from regression and classification to NLP and computer vision), and capable of forecasting complex data patterns. 
Our main contributions are therefore threefold:
\begin{itemize}
    \item To achieve 
    the model compactness, 
    we removed channel expansion, used fixed-size projections, applied gating via elementwise multiplication, and replaced multi-head attention with single-layer Conv1D filters.
    \item To enable cross-domain generalization, the architecture employs task-specific input embeddings—using direct numerical mappings for regression and classification, token and positional encodings for sequential patterns in NLP, and convolutional layers for spatial hierarchies in vision—while preserving a fixed, shared core representation across domains.
    \item To achieve effective complex pattern detection with low model complexity, we used gated residual paths to retain weak signals, 
    ReLU
    to maintain gradient stability across varying activations, and Conv1D as a bandpass filter to extract local dependencies.
\end{itemize}
\section{Related Work}
To address the challenges of resource-constrained environments, numerous lightweight models have been developed for machine learning tasks in data. 
However, traditional models often fail to balance computational efficiency with accurate and robust pattern recognition. These limitations become particularly evident in on-device AI applications that require high accuracy despite limited resources \cite{thota2024optimizing}. 
There are numerous architectures to build such models, namely, 
feed-forward, recurrent, CNN, and TinyML architecture.

\subsection{Feed-Forward Architectures}
Feed-forward architectures offer simplicity and computational efficiency through unidirectional information flow without feedback. They are well-suited for on-device AI due to their support for parallel processing and low inference latency. However, their lack of memory, 
limits their effectiveness on sequential tasks \cite{dhar2021survey}. 

To address the feed-forward limitations,
several models based on 
MLPs
have been proposed. One such model is the gMLP, which introduces a spatial gating mechanism to facilitate interactions across input dimensions while preserving architectural simplicity \cite{NEURIPS2021_4cc05b35}. Its simple structure allows for fast inference and reduced memory usage, which makes it suitable for edge devices. Nonetheless, the absence of explicit temporal modeling impairs its ability to learn long-range dependencies. For instance, in activity recognition, where input data arrive as continuous time-series, gMLP’s purely feed-forward design may hinder its ability to capture transitions between activities that potentially degrade performance in real-world scenarios.

Another recent MLP-based model, the 
TSMixer,
captures complex temporal patterns by integrating information across time steps and feature dimensions \cite{chen2023tsmixer}. Despite its effectiveness, TSMixer requires a relatively large number of parameters and significant computational resources, which poses challenges for processing multivariate sensor data in wearable devices. In addition to that, its structural complexity limits real-time, energy-efficient processing on resource-constrained devices such as smartphones and smartwatches. To support latency-sensitive applications, further optimization and lightweight adaptations may be necessary to preserve its predictive capabilities \cite{chen2023tsmixer}.


\subsection{Recurrent Architectures}
Unlike the feed-forward architecture, which has unidirectional information flow without any feedback, the recurrent architecture features cyclical connections, maintains an internal state, and is well-suited for processing sequential data. They effectively capture temporal dependencies and contextual information; which make them valuable for tasks such as time-series analysis, natural language processing, and speech recognition. However, default recurrent architectures have an inherently sequential nature that hinders parallel computation. This results in higher latency and increased energy consumption, both of which are critical concerns for real-time and resource-constrained environments. To address these challenges and make recurrent architectures suitable for on-device deployment, various adaptations and optimizations have been proposed. 
The most widely adopted architecture for time series data is the long short-term memory (LSTM) network, which has demonstrated exceptional performance in sequential data processing by effectively capturing long-term dependencies through gating mechanisms. However, LSTM based models typically require a large number of parameters and prolonged training periods, which leads to high computational complexity and memory overhead \cite{hochreiter1997long}. Moreover, when applied to long sequences, LSTM-based models are prone to overfitting, which degrades generalization performance \cite{karpathy2015visualizing}. In order to optimize these architectures, it often requires additional computational costs, which further limit their scalability and suitability for latency-sensitive applications \cite{pmlr-v37-jozefowicz15}. Owing to these reasons, the LSTM-based approaches are less viable for deployment in resource-constrained environments such as mobile devices, embedded systems, and edge computing platforms.
To overcome the limitations of LSTM, the gated recurrent unit (GRU) architecture was introduced, where the gating mechanism was simplified by reducing the three gates (input, forget, and output) to two: reset and update gates.
This 
design results in approximately 25\% fewer parameters and a lower memory consumption during training and inference \cite{cho-etal-2014-learning}. Despite these advantages, GRU architecture still requires significant computational resources when processing long sequences due to its inherently sequential nature.

        \begin{figure*}[t]
            \includegraphics[width=17.5cm]{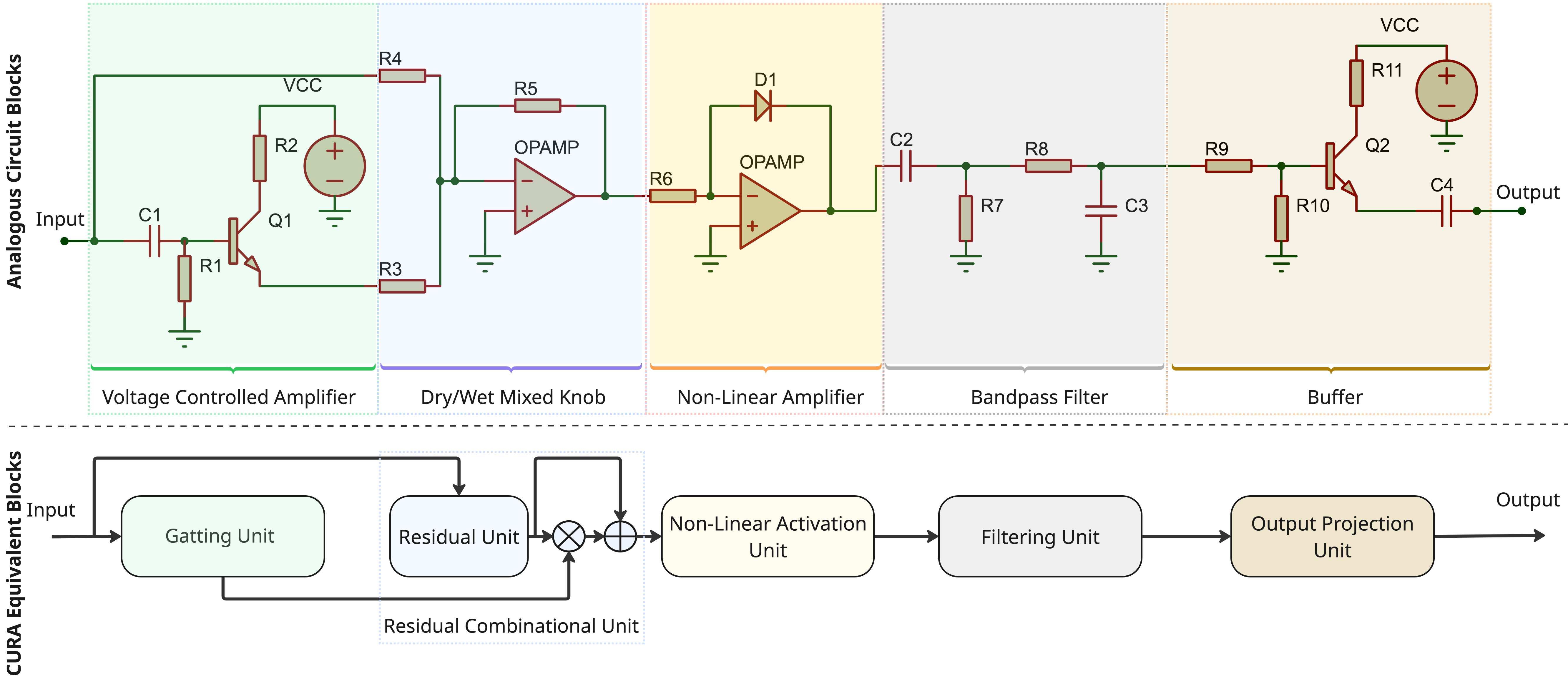}
            \centering
            \caption{Audio signal processing circuit and equivalent block components of \ours{}.} 
            \label{circuit}
        \end{figure*}

\subsection{Lightweight CNN Architectures}
In contrast to recurrent architectures, which are inherently sequential, and feed-forward models adapted for sequence tasks, convolutional neural networks (CNNs) are purpose-built for lightweight modeling in vision-centric applications. Their strength lies in efficiently extracting hierarchical features from structured data such as images and sensor grids.
Recent advances in lightweight CNNs have optimized architectural components to balance accuracy, latency, and resource usage, producing compact yet efficient models suitable for mobile and embedded deployment \cite{sandler2018mobilenetv2}.
However, CNNs are primarily optimized for visual data and often struggle to generalize in broader tasks such as time-series forecasting, natural language understanding, and multimodal processing. Despite their original computational demands, modern lightweight CNNs remain effective for vision tasks in resource-constrained environments.

One of the widely used CNN architecture that achieves a favorable trade-off between accuracy and computational efficiency is EfficientNet~\cite{tan2019efficientnet}. 
EfficientNet introduces a compound scaling strategy that uniformly scales network depth, width, and input resolution to improve performance without incurring excessive resource costs. 
Similarly, MobileNetV3, designed via neural architecture search, is optimized for high performance under stringent latency and power constraints~\cite{howard2019searching}.
This architecture incorporates efficient building blocks such as squeeze-and-excitation modules and hard-swish activation functions to support fast inference on mobile and embedded platforms. Both EfficientNet and MobileNetV3 have demonstrated strong performance on image classification benchmarks such as ImageNet and are widely adopted in resource-aware visual recognition tasks
\cite{tan2019efficientnet}.

Unfortunately, the structural rigidity of CNN-based models poses a significant challenge when applying them to non-visual domains. Specifically, in tasks like time-series forecasting, natural language processing, or multivariate analysis, CNN architectures often lack the crucial inductive biases and temporal modeling capabilities necessary for robust pattern recognition~\cite{chen2023tsmixer}. 

\subsection{TinyML Architectures}

TinyML denotes the specialized field concerned with the deployment of machine learning models on highly resource-constrained edge devices, primarily microcontroller units. 
These devices are characterized by severe limitations in memory, processing power, and energy budgets \cite{warden2019tinyml, capogrosso2024machine}. Models developed within the TinyML paradigm are inherently ultra-lightweight, often constrained to mere tens of kilobytes in size, and are rigorously optimized for real-time inference coupled with minimal energy consumption.


TinyML models are highly efficient, bringing intelligent capabilities to power-constrained edge devices like wearables, IoT sensors, and industrial monitoring systems. However, a significant limitation of these solutions is their inherently task-specific nature. They often rely on highly simplified architectures, which inherently compromises their expressive capacity and generalization performance \cite{shi2020communication}.
Due to this task-specific design and architectural simplicity, many TinyML solutions struggle considerably with complex input distributions or generalizability across diverse domains. For instance, using these models for tasks like predicting multiple time-series variables, classifying text data, or combining different types of data (like text and images) remains quite challenging \cite{chen2023tsmixer, nosouhian2021review, yamak2019comparison}.
Despite advancements in lightweight model design, a fundamental trade-off persists between model size and representational capacity across existing architectures, including feed-forward networks, recurrent neural networks, CNN-based models, and TinyML approaches. 
When these models are scaled down to run on devices with limited resources, they often struggle to understand complex patterns or recognize detailed features.
%
There are techniques like quantization, pruning, and knowledge distillation that offer partial solutions for model compression or knowledge transfer from larger networks; however, they frequently fall short of preserving the expressive power required for high-accuracy inference in complex, real-world scenarios. Consequently, these existing solutions struggle to generalize across diverse tasks or effectively process complex, multimodal inputs in resource-constrained environments.
%



\section{Proposed Architecture: CURA} \label{sec: proposed architecture}
In light of the limitations observed across existing 
architectures there is a growing demand for models that strike a more effective balance between computational efficiency and representational power. 
To address this gap, we introduce \ours{}, a “simple yet powerful” architecture tailored for on-device AI. 
\ours{} is designed to integrate local and global information through a structurally efficient architecture. It emphasizes interpretability\footnote{Interpretability here refers to architectural transparency and modularity, which allow for easier inspection of information flow and component-level behavior, rather than formal explainability or post-hoc analysis.} 
and functional compactness for a robust and scalable solution in resource-constrained environments.

\subsection{Architectural Overview and Circuit Analogy}
\ours{} is inspired by an analog audio signal processing circuit shown in Figure \ref{circuit} (with its equivalence blocks).
The core design philosophy of \ours{} translates the functional roles of these analog circuit components into neural network modules.
In the analog circuit, the processing flow is as follows:
\begin{enumerate}
    \item An input signal (e.g., an audio waveform) is first fed into the VCA. The VCA dynamically modulates the signal's amplitude in response to an external control voltage, which is crucial for shaping envelopes and controlling signal dynamics in synthesis.
    \item The modulated signal then proceeds to a dry/wet mixer. This stage combines the original (dry) input signal with the processed (wet) signal from the VCA, enabling a precise adjustment of the effect strength to balance the original signal characteristics with the desired processing intensity.
    \item The mixed output is routed to a non-linear amplifier. This component introduces harmonic content, tonal coloration, and sonic character through controlled distortion or saturation, which enriches the sound by adding harmonic complexity and a warmer or more aggressive quality.
    \item The signal then passes through a bandpass filter. This filter selectively focuses on a specific frequency range, attenuating unwanted low- and high-frequency content. It isolates musically relevant portions of the spectrum and eliminates artifacts from previous stages.
    \item Finally, a buffer stage is employed. This output stage prevents signal degradation and ensures proper impedance matching between circuit stages, guaranteeing clean, stable signal transmission without introducing tone coloration or loading effects that could compromise audio quality.
\end{enumerate}

Inspired by this signal processing circuit, the proposed \ours{} architecture also comprises five main computational components, where each component corresponds to the aforementioned analog circuit blocks (See Figure \ref{circuit}).  
In the following, details of each of the blocks are presented.
%


\vspace{0.2cm}
\noindent\textbf{1. Gating unit.}
Let the input observations be $\bm{X} \in \mathbb{R}^{L \times C}$, where $L$ represents the sequence length and $C$ denotes the number of input variables.
The gating path within CURA functions as a dynamic input modulator, analogous to the VCA in analog circuits. This component adaptively controls the flow of input information through the network by generating a modulation signal 
\begin{equation}
\bm{g} = \sigma \left( \bm{X}\bm{W}_g + \bm{b}_g \right),
\label{eq:gating_path_def}
\end{equation}
where $\sigma$ is an activation function, $\bm{W}_g$ represents the learnable gating weights, and $\bm{b}_g$ is the gating bias. The gating path dynamically adjusts the signal based on learned contextual cues. 
By selectively enhancing or suppressing signal features, it improves the processing pipeline's sensitivity to temporal or contextual variations while maintaining a low parameter count.


\vspace{0.2cm}
\noindent\textbf{2. Residual combinational unit.}
The second component of the architecture is the residual combinational unit, which combines two parallel transformations of the input $\bm{X}$: a gating projection defined in (\ref{eq:gating_path_def}), and a residual projection given by
\begin{equation}
    \bm{r} = \bm{X} \bm{W}_r + \bm{b}_r,
    \label{eq:residual_path_def}
\end{equation}
where $\bm{W}_r$ and $\bm{b}_r$ represent the weights and bias of the residual pathway. The output of the residual gated unit is then computed as:
\begin{equation}
    \bm{h}_1 = \bm{g} \odot \bm{r} + \bm{r},
    \label{eq:residual_gate}
\end{equation}
where \( \odot \) denotes the Hadamard product. Notice that (\ref{eq:residual_gate}) can be equivalently expressed as
\begin{equation}
    \bm{h}_1 = \bm{r} \odot \left( \bm{g} + \bm{1}\right),
    \label{eq:residual_gate_equivalent}
\end{equation}
which emphasizes the modulation of the residual signal by an additive gating factor. This formulation ensures that the base residual signal is always preserved while being adaptively scaled. This design facilitates stable gradient propagation, retains essential input features through direct pathways, and introduces a learnable amplification mechanism that enables selective feature enhancement up to a factor of 2.

\vspace{0.2cm}
\noindent\textbf{3. Nonlinear activation unit.}
The third unit, i.e., the nonlinear activation unit, is defined as
\begin{equation}
    \bm{h}_2 = \phi \left( \bm{h}_1 \bm{W}_n + \bm{b}_n \right),
    \label{eq:nonlinear_activation}
\end{equation}
where \( \phi(\cdot) \) denotes a smooth nonlinear activation function, such as ReLU, Tanh, or GELU. The parameters \( \bm{W}_n \) and \( \bm{b}_n \) represent the weights and bias of this unit.
The nonlinear activation unit in \ours{} introduces differentiable nonlinearity into the feature stream, analogous to a nonlinear amplifier in analog circuits. It enables the model to dynamically reshape signal characteristics, enhancing representational capacity through learned transformations. The use of smooth activations mitigates the abrupt gradient discontinuities of traditional ReLU, facilitating more stable optimization and richer gradient flow.

\vspace{0.2cm}
\noindent\textbf{4. Filtering unit.}
The filtering unit applies a filtering operation to the output of the nonlinear activation unit in (\ref{eq:nonlinear_activation}), represented as
\begin{equation}
    \bm{h}_3 = \mathcal{F}(\bm{h}_2; \Theta_f),
    \label{eq:filtering_unit}
\end{equation}
where \(\mathcal{F}(\cdot)\) denotes a filtering function parameterized by \(\Theta_f\), which can correspond to various types of filters such as convolutional, 
linear,
or attention-based mechanisms. 
This unit acts as a local feature extractor, capturing short-range dependencies and emphasizing relevant patterns within the feature stream. When instantiated as a convolutional filter, it is analogous to a bandpass filter in analog signal processing.




\vspace{0.2cm}
\noindent\textbf{5. Output projection unit.}
The final output of the architecture $\bm{y}$ is obtained via a linear projection of the filtered feature representation as
\begin{equation}
    \bm{y} = \bm{h}_3 \bm{W}_o + \bm{b}_o,
    \label{eq:output_projection}
\end{equation}
where \( \bm{W}_o  \) and \( \bm{b}_o \) are the weights and bias of the output layer. 
The output projection unit functions similarly to a buffer stage in analog circuits, which ensures the signal is impedance-matched and stabilized for clean delivery without degradation or tonal coloration. In \ours{}, the output projection unit maps the internal representation learned to task-specific targets, such as class probabilities or regression values, while preserving the integrity and scaling of the feature information.

The proposed \ours{} architecture is an ultra-lightweight MLP-based core structure designed to achieve high expressiveness and interpretability with an extremely small parameter count. In contrast to conventional deep learning architectures that typically employ repetitive and accumulative modules, \ours{} comprises independent and self-contained computational cores, each capable of performing specialized functions that extend beyond the role of traditional neural network blocks. This will be demonstrated in the following section.
\begin{table*}[!t]
    \centering
    \caption{Dataset Characteristics and Evaluation Scope}
    \begin{tabular}{p{2cm} p{2cm} p{1.8cm} p{1cm} p{2cm} p{2cm} p{4.2cm}}
        \toprule
        \multicolumn{7}{l}{\textbf{normal dataset}} \\
        \toprule
        \textbf{Dataset} & \textbf{Domain} & \textbf{Task Type} & \textbf{Samples} & \textbf{Features} & \textbf{Temporal Res.} & \textbf{Key Characteristics} \\
        \midrule
        UCI HAR & Human Activity & Classification & 10,299 & 561 & 50Hz (20ms) & Smartphone sensors, 6 classes \\
        ETTm1 & Power Systems & Regression & 69,680 & 7 & 15 minutes & Long-term forecasting \\
        FallAllD & Healthcare & Classification & 26,420 & 12 & High-freq. & Multi-sensor fusion \\
        House Prices & Tabular Data & Regression & 1,460 & 79 & N/A & House price prediction \\
        S\&P 500 & Finance & Regression & $\sim$3,650 & 5 & Daily & Stock market prices \\
        \toprule
        \multicolumn{7}{l}{\textbf{Cross-domain Datasets}} \\
        \toprule
        \textbf{Dataset} & \textbf{Domain} & \textbf{Task Type} & \textbf{Samples} & \textbf{Features} & \textbf{Temporal Res.} & \textbf{Key Characteristics} \\
        \midrule
        CIFAR-10 & Computer Vision & Classification & 60,000 & 3 (RGB) & N/A & 10 image classes \\
        AG News & NLP & Classification & 127,600 & 2 (title, desc.) & N/A & 4 news topics \\
        Amazon Polarity & NLP & Classification & 4,000,000 & 2 (title, body) & N/A & Sentiment (positive/negative) \\
        BoolQ & NLP & QA (Yes/No) & 16,000 & Passage + Ques. & N/A & Reading comprehension \\
        HellaSwag & NLP & NLI (Inference) & 70,000 & Context + Ending & N/A & Common sense inference \\
        \bottomrule
    \end{tabular}
    \label{tab:datasets}
\end{table*}

\section{Evaluation of CURA}

To rigorously evaluate the compactness, generalizability, and complex pattern forecasting of \ours{}, we conduct experiments across two distinct categories of datasets. 
The first includes five real-world datasets covering a wide range of multivariate time series classification and regression tasks. 
The second comprises cross-domain datasets 
to assess the performance of \ours{} beyond conventional time series settings, thereby testing its adaptability to varying domains, temporal granularities, sensor modalities, and task complexities.
This evaluation framework spans applications such as human activity recognition and financial market forecasting, demonstrating the effectiveness of \ours{} across diverse sequence lengths, feature dimensions, and learning objectives. 
A summary of the characteristics of the datasets is provided in Table~\ref{tab:datasets}, with detailed descriptions presented below.

\vspace{0.3cm}
\subsubsection{Time-series Datasets} 
We consider the following multivariate time series datasets. 


\vspace{0.3cm}
\noindent\textbf{Human activity recognition dataset:}
The UCI human activity recognition (UCI HAR) dataset \cite{anguita2013public} is a publicly available dataset created from recordings of 30 volunteers (aged 19-48 years) performing six different activities of daily living while carrying a waist-mounted samsung galaxy S-II smartphone.
The dataset captures human activities including walking, walking upstairs, walking downstairs, standing, sitting, and laying down using the smartphone's built-in accelerometer and gyroscope sensors. The data was collected in laboratory conditions but volunteers were asked to perform the activities freely for a more naturalistic dataset, with each subject performing the protocol twice under different smartphone positioning conditions. The dataset 
serves as a benchmark for smartphone-based human activity recognition research.

\vspace{0.3cm}
\noindent\textbf{Electricity transformer temperature dataset:}
The electricity transformer temperature (ETT) monitoring dataset is a high-resolution multivariate time-series dataset designed for long-sequence forecasting in power systems \cite{Zhou_Zhang_Peng_Zhang_Li_Xiong_Zhang_2021}. 
We use the ETTm1 variant, which consists of 15-minute sampled data collected over one year from a transformer station in China. 
Each data point includes the target variable “oil temperature” and six associated features representing power load metrics across different voltage levels. 
The dataset is partitioned into 12 months for training, 4 months for validation, and 4 months for testing, enabling a temporally consistent evaluation protocol.


\vspace{0.3cm}
\noindent\textbf{FallAllD dataset:}
FallAllD is a publicly available time-series dataset comprising 26,420 recordings collected from 15 participants performing simulated falls and daily activities~\cite{saleh10fallalld}. 
The data were captured using sensors worn on the waist, wrist, and neck, including accelerometer, gyroscope, magnetometer, and barometer measurements. 
For this study, the dataset is used for binary classification to distinguish fall events from non-fall activities, supporting evaluation of human activity recognition systems in wearable sensing scenarios.


\vspace{0.3cm}
\noindent\textbf{House price dataset:}
The Kaggle house prices dataset consists of 1,460 residential property records from Ames, Iowa, each annotated with the final sale price~\cite{houseprices}. 
It includes 79 explanatory features describing various property characteristics, such as size, quality, location, and construction details. 
This dataset is used for regression, with the objective of predicting the target variable sale price based on the provided features.


\vspace{0.3cm}
\noindent\textbf{S\&P 500 dataset (2010–2024):}
This dataset consists of historical daily stock data for companies listed in the S\&P 500 index from 2010 to 2024, retrieved via the Yahoo! finance API~\cite{aroussi2025yfinance}. 
Each entry includes open, high, low, close, adjusted close prices, and trading volume. 
The dataset is used for regression tasks focused on predicting future stock prices. 
Its long temporal span and high dimensionality provide a challenging benchmark for evaluating model performance in volatile, real-world financial conditions.

\subsubsection{Cross Domain Datasets} 
To verify the generalizability of CURA to cross-domain applications such as computer vision and natural language processing, we perform further evaluations on the following diverse datasets that span different modalities and problem types:


\vspace{0.3cm}
\noindent\textbf{CIFAR-10 dataset:}
The CIFAR-10 dataset is a well-established benchmark for image classification, consisting of 60,000 color images at a resolution of 32×32 pixels, categorized into 10 classes: airplane, automobile, bird, cat, deer, dog, frog, horse, ship, and truck~\cite{krizhevsky2009learning}. Each class contains 6,000 images, with a standard split of 50,000 for training and 10,000 for testing. 


\vspace{0.3cm}
\noindent\textbf{AG News dataset:}
The AG News dataset is a widely used benchmark for topic classification, compiled from over 2,000 news sources \cite{NIPS2015_250cf8b5}. It comprises 120,000 training samples and 7,600 test samples evenly distributed across four categories: world, sports, business, and science/technology. Each entry includes a news title and a short description, making it well-suited for evaluating text classification models.


\vspace{0.3cm}
\noindent\textbf{Amazon polarity dataset:}
The amazon polarity dataset is a large-scale benchmark for binary sentiment classification, derived from Amazon product reviews \cite{NIPS2015_250cf8b5}. It contains 3.6 million training samples and 400,000 test samples, each labeled as positive or negative based on user star ratings. Reviews include both titles and full text bodies for evaluating text-based binary sentiment classification models.


\vspace{0.3cm}
\noindent\textbf{BoolQ dataset:}
BoolQ is a reading comprehension dataset comprising 16,000 naturally occurring yes/no questions paired with supporting passages from Wikipedia~\cite{clark-etal-2019-boolq}. The questions originate from real Google search queries and are annotated with binary answers ("yes" or "no"). Due to the need for contextual reasoning and inference, BoolQ poses a significant challenge and is used as a benchmark for evaluating natural language understanding models.

\begin{table*}[!ht]
    \centering
    \caption{Ablation Study of \ours{} Across Architectural Design Choices}
    \renewcommand{\arraystretch}{1.2}
    \begin{tabular}{p{3.5cm} p{3.5cm} p{5.2cm} p{2.2cm}}
        \toprule
        \textbf{Variation Axis} & \textbf{Design Component} & \textbf{Ablation Variant} & \textbf{F1 Score (\%)} \\
        \midrule
        \midrule
        
        \multirow{3}{*}{Gating mechanism}
            & \multirow{3}{*}{Gating unit}
                & Multiplicative (default) & 94.70 \\
            & 
                & Linear & 94.34 \\
            & 
                & Convolutional & 94.20 \\
        \midrule
        
        \multirow{5}{*}{Activation function}
            & \multirow{2}{*}{Gating unit} 
                & Sigmoid (default) & 94.70 \\
            & 
                & Hard sigmoid & 91.00 \\
        \cmidrule(l){2-4}
            & \multirow{3}{*}{Nonlinear activation unit} 
                & 
                ReLu (default)
                & 
                94.70 \\
            & 
                & GELU & 94.34 \\
            & 
                & Tanh + Conv1D & 94.20 \\
        \midrule
        
        \multirow{3}{*}{Filtering module}
            & \multirow{3}{*}{Filtering unit}
                & Conv1D (default) & 94.70 \\
            & 
                & Linear filter (1×1 Conv) & 90.00 \\
            & 
                & No filter & 90.00 \\
        \bottomrule
    \end{tabular}
    \label{tab:ablation-merged}
\end{table*}


\vspace{0.3cm}
\noindent\textbf{HellaSwag dataset:}
HellaSwag is a benchmark for commonsense natural language inference with longer, diverse contexts from ActivityNet and WikiHow~\cite{zellers-etal-2019-hellaswag}. This dataset was constructed using adversarial filtering, it includes machine-generated endings carefully refined to challenge and mislead models. The dataset evaluates a model’s ability to plausibly complete event descriptions in complex, realistic scenarios.

\vspace{0.3cm}
Based on the aforementioned datasets, we evaluate different aspects of \ours{} to comprehensively validate its effectiveness and practical applicability. 
We begin the evaluation by conducting an ablation study, where we systematically remove or replace with alternative an individual components of the architecture to assess their contribution to overall performance and identify the most critical design elements. 
Next, we assess the core objectives of the proposed architecture through multiple evaluation criteria. 
First, we evaluate high expressiveness, which measures how effectively our architecture can detect and predict complex patterns in datasets with minimal computational complexity, demonstrating its ability to capture intricate temporal dependencies and multivariate relationships. 
%
%
Second, we evaluate parameter efficiency by comparing total parameter counts with established baselines, showing that \ours{} achieves 
comparable or
superior performance with significantly fewer parameters, making it suitable for resource-constrained environments.
Finally, we examine the interoperability and generalizability of \ours{} by evaluating how our architecture adapts to diverse application domains beyond time series analysis, including computer vision tasks (image classification, object detection), natural language processing applications (text classification, sequence modeling), and other structured data problems, thereby validating its versatility as a general-purpose deep learning architecture.

\subsection{Structural Ablation Study of CURA}

As detailed in Section \ref{sec: proposed architecture}, \ours{} comprises five key units: 
(1) gating path, 
(2) residual gated unit, 
(3) nonlinear activation unit, 
(4) filtering unit, and 
(5) output projection unit.
To evaluate the contribution of each component to overall performance, we conducted an ablation study on the UCI HAR dataset. 
Table~\ref{tab:ablation-merged} shows the F1 scores resulting from systematic variations along three architectural axes: gating mechanisms, activation functions, and filtering modules.
Based on extensive preliminary experiments, the default \ours{} configuration employs multiplicative gating, ReLU activation, and a Conv1D filtering module. This configuration achieves an F1 score of 94.70\%.
Subsequent rows show the effect of replacing each default component with alternative variants, highlighting their relative impact on performance.

\subsubsection{Gating Mechanism}

To validate the default choice of multiplicative-based gating, we systematically replaced it with alternative mechanisms: linear gating and convolutional (Conv1D) gating. These substitutions led to performance degradation of 0.38\% and 0.53\%, respectively.
This empirical evidence indicates that multiplication gating in \ours{} effectively regulates information flow while preserving signal fidelity and parameter efficiency. Its superior performance likely arises from its alignment with the residual modulation behavior intrinsic to \ours{}'s architecture, where multiplicative operations facilitate stable gradient flow and maintain feature magnitude consistency.
In contrast, linear gating introduces additional parameters via weight matrices and may create gradient propagation bottlenecks. Convolutional gating, while more flexible, incurs higher computational cost and imposes spatial dependencies that can disrupt the intended pointwise feature scaling. The efficacy of elementwise product gating stems from its simplicity and direct mapping to residual modulation, where each feature dimension is independently scaled without added cross-dimensional interactions or parameter overhead.

\vspace{0.3cm}
\subsubsection{Activation Function}
\ours{} incorporates activation functions in two key components: the gating path and the nonlinear activation unit.
In the gating path, the default activation is the Sigmoid function, which facilitates smooth gradient flow and stable optimization. Replacing it with a Hard Sigmoid results in a 3.70\% drop in F1 score (Table~\ref{tab:ablation-merged}), primarily due to the Hard Sigmoid’s saturation regions and non-smooth gradients, which hinder effective learning and reduce sensitivity to nuanced input variations.
In the nonlinear activation unit, we evaluated GELU and Tanh+Conv1D as alternatives to the default Soft-ReLU. Both alternatives resulted in modest performance declines (F1 = 94.34\% and 94.20\%, respectively), suggesting that ReLU achieves a better balance between representational power, activation sparsity, and gradient stability. While GELU offers smooth probabilistic gating, it may blur critical feature boundaries. Tanh+Conv1D enhances local pattern extraction but introduces saturation effects and additional computational cost without corresponding performance improvements.

\begin{table*}[!ht]
    \centering
    \caption{Performance and Parameter Efficiency Comparison of \ours{} and Baselines Across Multiple Datasets}

    \begin{tabular}{p{2.5cm} p{2.5cm} p{2cm} p{2cm} p{2cm} p{2cm} p{2cm}}
        \toprule
        \textbf{Dataset} & \textbf{Metric} & \textbf{CURA} & \textbf{gMLP\cite{NEURIPS2021_4cc05b35}} & \textbf{GRU\cite{nosouhian2021review}} & \textbf{LSTM\cite{hochreiter1997long}} & \textbf{TSMixer\cite{chen2023tsmixer}} \\
        \midrule\midrule
        \multirow{3}{*}{S\&P 500} 
            & No. of Params. & 746 & 1,067 & 4,478 & 4,513 & 3,585 \\
            & \( R^2 \) Score (\%) & 99.0 & 99.0 & 97.0 & 98.0 & 99.5 \\
            & Parameter Efficiency & 0.130 & 0.092 & 0.021 & 0.021 & 0.027 \\
        \midrule
        \multirow{3}{*}{House Prices} 
            & No. of Params. & 790 & 929 & 154,561 & 1,285 & 11,475 \\
            & \( R^2 \) Score (\%) & 84.0 & 79.0 & -0.12 & 21.0 & 71.0 \\
            & Parameter Efficiency & 0.10 & 0.085 & -7.76 & 0.016 & 0.0061 \\
        \midrule
        \multirow{3}{*}{ETTm1} 
                & No. of Params. & 731 & 11,423 & 14,081 & 18,753 & 55,090 \\
            & \( R^2 \) Score (\%) & 86.45 & 80.88 & 53.70 & 44.46 & 91.15 \\
            & Parameter Efficiency & 0.12 & 0.0070 & 0.0038 & 0.0023 & 0.0016 \\
        \midrule
        \multirow{3}{*}{UCI HAR} 
            & No. of Params. & 2,342 & 2,374 & 15,174 & 20,102 & 14,490 \\
            & F1 Score (\%) & 95.40 & 94.96 & 93.16 & 93.71 & 95.63 \\
            & Parameter Efficiency & 0.041 & 0.040 & 0.0064 & 0.0046 & 0.0065 \\
        \midrule
        \multirow{3}{*}{FallAllD} 
            & No. of Params. & 345 & 440,551 & 7,681 & 4,769 & 836,571 \\
            & F1 Score (\%) & 80.0 & 79.3 & 77.5 & 77.3 & 76.1 \\
            & Parameter Efficiency & 0.231 & 0.00018 & 0.01 & 0.016 & 0.000091 \\
        \bottomrule
    \end{tabular}
    \label{tab:unified-model-results}
\end{table*}

\vspace{0.3cm}
\subsubsection{Filtering Module}
%
\ours{} employs 1D convolutional filter to capture local temporal dependencies within the feature stream. To assess its effectiveness, we evaluated two alternatives: removing the filter entirely and replacing it with a linear layer (equivalent to a $1 \times 1$ convolution). In both cases, the F1 score dropped to 90\%, a decrease of 4.7\% 
compared to the default configuration, as shown in Table~\ref{tab:ablation-merged}.  
This performance degradation highlights the critical role of 1D convolution in modeling short-range temporal correlations and providing translation invariance—capabilities that are essential for multivariate time series classification. Without any filtering, the model fails to capture meaningful sequential patterns, treating each timestep in isolation. Similarly, the linear filter lacks a receptive field and operates as a point-wise transformation, rendering it incapable of extracting temporal dependencies across multiple steps. These results highlights the necessity of localized temporal feature extraction.

\subsection{Assessment of Model Compactness}

To evaluate the compactness and efficiency of \ours{} and baselines, we define the parameter efficiency metric as
\begin{equation}
    \eta = \frac{M}{P},
\end{equation}
where \( M \) denotes the model’s predictive performance and \( P \) is the number of trainable parameters. This metric offers a normalized view of performance per parameter, facilitating fair comparisons among models with different complexities. A higher value of \( \eta \) indicates a more efficient use of parameters, which is particularly important in resource-constrained environments. In our evaluation, \( M \) corresponds to the coefficient of determination (\( R^2 \)) for regression tasks and the F1 score for classification tasks.
To assess parameter efficiency, we evaluated \ours{} across all time-series datasets listed in Table~\ref{tab:datasets}, in comparison with SOTA architectures, including gMLP, GRU, LSTM, and TSMixer. The S\&P 500, house prices, and ETTm1 datasets were used for regression tasks involving continuous targets, while the UCI HAR and FallAllD datasets were employed for classification tasks with discrete categorical labels.
To ensure fair and consistent evaluation, all models were trained under identical experimental settings.
We first conduct time-series forecasting experiments on the S\&P 500 dataset, evaluating all models using the \( R^2 \) score. A fixed input window of 20 time steps was used, while the number of input channels was varied from 1 to 4 to assess model scalability across different feature dimensions and multivariate configurations.
As shown in Table~\ref{tab:unified-model-results}, \ours{} achieves an exceptional \( R^2 \) score of 99\% using only 746 parameters, which significantly outperforms all baseline models in terms of parameter efficiency. Although gMLP reaches a similar level of predictive accuracy, it requires considerably more parameters. The efficiency gap becomes more pronounced with recurrent architectures: GRU and LSTM requiring approximately six times more parameters to achieve comparable performance.
%
The TSMixer slightly outperforms \ours{}, however, it does so at the cost of nearly five times more parameters, which highlights a less favorable trade-off between accuracy and model compactness.
For the house price data set, \ours{} achieves an \( R^2 \) score of 84\% with only 790 parameters (which results in a parameter efficiency of 0.106). In contrast, all baseline models require significantly more parameters to achieve similar or lower performance. For example, the closest competitor, gMLP, uses 18\% more parameters (929 vs. 790) while achieving a score of 5 percentage points lower \( R^2 \) (i.e., 79\% vs. 84\%). The GRU and LSTM exhibit substantially degraded performance, with GRU producing a negative prediction capability (\( R^2 = -2\% \)) and LSTM reaching only 21\% accuracy, despite both employing over 1,200 parameters. The TSMixer achieves a moderate \( R^2 \) score of 71\%, yet this comes at the cost of 11,475 parameters, leading to a poor parameter efficiency of just 0.0061.



\begin{table*}[!t]
    \centering
    \caption{CURA Configuration Details for Various NLP Tasks}
    \begin{tabular}{p{4cm} p{5.5cm} p{7cm}}
        \toprule
        \multicolumn{3}{l}{\textbf{4-way Multiple-Choice Reasoning (HellaSwag)
        }} \\
        \toprule
        \textbf{Hyperparameter} & \textbf{Value} & \textbf{Description} \\
        \midrule
        Tokenizer & DistilBERT (\texttt{distilbert-base-uncased}) & All experiments use uncased DistilBERT tokenizer \\
        \addlinespace
        Input Format & Passage + 4 Choices & MCQ input format with context and alternatives \\
        \addlinespace
        Embedding Dimension & 192 & Token embedding dimension \\
        \addlinespace
        Hidden Dimension & 270 & Intermediate hidden size within CURA core \\
        \addlinespace
        Number of CURA Cores (Blocks) & 6 & Number of stacked attention-based blocks \\
        \addlinespace
        Number of Attention Heads & 6 & Number of parallel attention branches \\
        \addlinespace
        Final Pooling & AttentionPooling & Learns weighted summary over token sequence \\
        \addlinespace
        Output Layer & Linear(192 $\rightarrow$ 1) & Produces a single score per answer choice \\
        \addlinespace
        Number of Output Classes & 4 & One logit per candidate choice \\        
        \midrule
        \multicolumn{3}{l}{\textbf{Multi-class Classification (AG News)}
        } \\
        \toprule
        Tokenizer & DistilBERT (\texttt{distilbert-base-uncased}) & All experiments use uncased DistilBERT tokenizer \\
        \addlinespace
        Input Format & Text & Raw news content used as input \\
        \addlinespace
        Embedding Dimension & 128 & Token embedding dimension \\
        \addlinespace
        Hidden Dimension & 180 & Internal hidden projection size \\
        \addlinespace
        Number of CURA Cores (Blocks) & 3 & Number of sequential attention blocks \\
        \addlinespace
        Number of Attention Heads & 2 & Independent self-attention branches \\
        \addlinespace
        Final Pooling & AttentionPooling & Learns importance-weighted token summary \\
        \addlinespace
        Output Layer & Linear(128 $\rightarrow$ 4) & Outputs one logit per class \\
        \addlinespace
        Number of Output Classes & 4 & 4-class classification for topic labels \\
        \midrule
        \multicolumn{3}{l}{\textbf{Binary Sentiment Classification (Amazon Polarity)}
        } \\
        \toprule
        Tokenizer & DistilBERT (\texttt{distilbert-base-uncased}) & All experiments use uncased DistilBERT tokenizer \\
        \addlinespace
        Input Format & Text & Review content with positive/negative sentiment \\
        \addlinespace
        Embedding Dimension & 128 & Token embedding dimension \\
        \addlinespace
        Hidden Dimension & 180 & Internal hidden projection size \\
        \addlinespace
        Number of CURA Cores (Blocks) & 3 & Number of attention layers \\
        \addlinespace
        Number of Attention Heads & 2 & Self-attention branches \\
        \addlinespace
        Final Pooling & AttentionPooling & Learns token importance over sequence \\
        \addlinespace
        Output Layer & Linear(128 $\rightarrow$ 2) & Binary classifier output head \\
        \addlinespace
        Number of Output Classes & 2 & Positive vs. negative sentiment \\
        \midrule
        \multicolumn{3}{l}{\textbf{Contextual Binary QA (BoolQ)}
        } \\
        \toprule
        Tokenizer & DistilBERT (\texttt{distilbert-base-uncased}) & All experiments use uncased DistilBERT tokenizer \\
        \addlinespace
        Input Format & Question + Passage & QA format with Yes/No answer based on context \\
        \addlinespace
        Embedding Dimension & 128 & Dimension of input token embeddings \\
        \addlinespace
        Hidden Dimension & 150 & Feedforward projection inside each block \\
        \addlinespace
        Number of CURA Cores (Blocks) & 4 & Number of stacked attention layers \\
        \addlinespace
        Number of Attention Heads & 8 & Multi-head attention branches \\
        \addlinespace
        Final Pooling & AttentionPooling & Learns weighted sequence summary \\
        \addlinespace
        Output Layer & Linear(128 $\rightarrow$ 2) & Outputs Yes/No logit scores \\
        \addlinespace
        Number of Output Classes & 2 & Binary QA output \\
        \bottomrule
    \end{tabular}
    \label{tab:cura_nlp_config}
\end{table*}


On the ETTm1 dataset, \ours{} significantly outperforms gMLP and GRU in terms of $R^2$ score while maintaining a much smaller model size. Specifically, \ours{} uses 15.6$\times$ fewer parameters than gMLP and 19.3$\times$ fewer than GRU. Although TSMixer achieves a slightly higher $R^2$ score, it does so with a model comprising over 55{,}000 parameters, in contrast to the compact architecture of \ours{}. These results highlight that \ours{} offers competitive accuracy while achieving the lowest model complexity among all evaluated methods.


\begin{table*}[!t]
    \centering
    \caption{Comparison of F1 Score and Parameter Count Between \ours{} and bart-base Across NLP Tasks}
    \renewcommand{\arraystretch}{1.2}
    \begin{tabular}{p{6cm} p{2.5cm} p{2.5cm} p{2.5cm} p{2.5cm}}
        \hline
        \multirow{2}{*}{\textbf{Task}} 
        & \multicolumn{2}{c}{\textbf{\ours{}(from scratch)}} 
        & \multicolumn{2}{c}{\textbf{bart-base\cite{lewis2019bart}(from scratch)}} \\
        \cline{2-5}
        & \textbf{F1 (\%)} & \textbf{No. of Params} & \textbf{F1 (\%)} & \textbf{No. of Params} \\
        \hline\hline
        4-way Multiple-Choice Reasoning & 28.79 & 3.26M & 32.37 & 140M \\ 
        \hline
        Multi-class Classification       & 88.39 & 1.81M & 38.03 & 140M \\
        \hline
        Binary Sentiment Classification & 90.69 & 1.81M & 27.22 & 140M \\
        \hline
        Contextual Binary QA            & 61.01 & 1.81M & 63.21 & 140M \\
        \hline
    \end{tabular}
    \label{tab:comparison-ours-bart-base}
\end{table*}

We also compared \ours{} with other baselines based on classification datasets, namely UCI HAR and FallAIID as shown in Table  \ref{tab:unified-model-results} under identical training settings, including optimizer, learning rate, learning rate scheduler, number of epochs, and batch size. To ensure a fair comparison, the
input format was standardized across models, either as a flattened vector 
or as a structured sequence, 
depending on the architectural requirements. 
As shown in Table~\ref{tab:unified-model-results}, \ours{} achieves F1 scores of approximately 95\% and 80\% on the UCI HAR and FallAIID datasets, respectively, using only 2.3K and 345 parameters. In comparison to its counterparts, including gMLP, GRU, LSTM, and TSMixer, our method demonstrates higher parameter efficiency while achieving comparable F1 scores. These results demonstrate that \ours{} provides a competitive tradeoff between accuracy and computational complexity.

\subsection{Assessment of Generalizability}
In addition to the aforementioned applications such as time-series regression and classification tasks, \ours{} demonstrates remarkable generalizability by being adopted for natural language understanding and vision tasks. 
This modular design highlights \ours{}'s potential as a domain-agnostic foundational architecture, which is capable of adapting to heterogeneous input modalities through targeted front-end modifications while preserving the computational efficiency and interpretability of its core structure.
This adaptability is achieved through a unified architectural approach where the core \ours{} structure remains consistent across domains while incorporating minimal, modality-specific front-end components to handle distinct types of input data. 

\subsubsection{Evaluation of NLP Tasks}
%
To handle NLP tasks, \ours{} integrates attention-based mechanisms in its front-end, specifically designed for sequential text processing. It employs low-rank token embeddings combined with rotary positional encoding to effectively model temporal dependencies in textual inputs. To demonstrate its versatility, we evaluate \ours{} across four representative NLP tasks: multiple-choice reasoning, multi-class news classification, binary sentiment analysis, and contextual question answering. The corresponding front-end configurations for each task are detailed in Table~\ref{tab:cura_nlp_config}. For performance benchmarking, we compare \ours{} against the BART-base architecture~\cite{lewis2019bart}, a strong 
baseline widely adopted in NLP applications.
Table~\ref{tab:comparison-ours-bart-base} presents a comparative evaluation of \ours{} and BART-base across several representative NLP tasks in terms of F1 score and parameter count. On the 4-way multiple-choice reasoning task (HellaSwag), \ours{} achieves an F1 score of 28.79\%, compared to 32.37\% by BART-base. However, this slight performance difference comes at a substantial computational cost: BART-base requires 140M parameters, approximately 
43$\times$
more than \ours{} (3.26M).
For multi-class and binary sentiment classification tasks, \ours{} significantly outperforms BART-base, achieving F1 scores of 88.39\% and 90.69\%, respectively, while BART-base yields only 38.03\% and 27.22\%, despite its considerably larger number of parameters. This highlights the strong parameter efficiency of \ours{} and its ability to deliver high performance with a lightweight architecture.
In the contextual binary QA task (BoolQ), \ours{} achieves an F1 score of 61.01\% compared to 63.21\% by BART-base, again using substantially fewer parameters.

These results demonstrate that \ours{} provides competitive, and in some cases superior, performance across a range of NLP tasks while requiring far fewer parameters, making it well-suited for deployment in resource-constrained environments.
It is worth noting that all results are obtained without any pretraining or fine-tuning. While such techniques are expected to improve the performance of both models, pretraining-based comparisons are beyond the scope of this study and are left for future work.

\begin{table*}[!t]
    \centering
    \caption{CURA Configuration Details for Vision Task (CIFAR-10)}
    \begin{tabular}{p{4cm} p{5.5cm} p{6.5cm}}
        \toprule
        \textbf{Hyperparameter} & \textbf{Value} & \textbf{Description} \\
        \midrule
        \midrule
        Task Type & 10-class Image Classification & For object recognition \\
        \addlinespace
        Input Format & $3 \times 32 \times 32$ RGB Image & Standard CIFAR-10 color image input \\
        \addlinespace
        CNN Feature Extractor & 
        3$\times$[Conv2d $\rightarrow$ BN $\rightarrow$ ReLU] $\rightarrow$ MaxPool2d(2), channels: 3$\rightarrow$64$\rightarrow$128$\rightarrow$256 
        & Three convolutional blocks, each consisting of two Conv-BN-ReLU layers followed by max pooling; channels increase at each stage. \\
        \addlinespace
        Flatten Dimension & 4096 & Output size after CNN and flattening ($256 \times 4 \times 4$) \\
        \addlinespace
        CURA Core Hidden Dimension & 32 & Hidden size inside CURA core (first Linear layer) \\
        \addlinespace
        CURA Core (Structure) & Gated Linear $\times$ Residual $\times$ Conv1D $\rightarrow$ LayerNorm $\rightarrow$ Dropout & Lightweight CURA block adapted for vision \\
        \addlinespace
        Output Layer & Linear(32 $\rightarrow$ 10) & Produces one logit per class \\
        \addlinespace
        Number of Output Classes & 10 & Total number of image categories \\
        \addlinespace
        Number of Parameters & $1.41$M & Total trainable parameters \\
        \addlinespace
        Optimizer & Adam (AMSGrad), weight decay = $1\times10^{-5}$ & Adaptive gradient descent with regularization \\
        \addlinespace
        Training Epochs & 130 & Total passes over the dataset \\
        \bottomrule
    \end{tabular}
    \label{tab:cura_vision_config}
\end{table*}

\subsubsection{Evaluation of Vision Tasks}
To configure \ours{} for vision tasks, we employed a shallow CNN consisting of two convolutional layers with batch normalization and ReLU activation, followed by max pooling. The hyperparameters for our vision configuration are detailed in Table~\ref{tab:cura_vision_config}. The resulting feature maps are flattened to 4096 dimensions and processed through a lightweight \ours{} core with a hidden dimension of 16. A post-convolutional projection layer expands the features to 64 channels before the final classification stage. The output layer maps the 64-dimensional features to 10 classification logits ($64 \rightarrow 10$), with global average pooling applied to reduce spatial dimensions. This vision variant contains approximately 266K parameters and was trained using the Adam optimizer with AMSGrad for 130 epochs.


Using the hyperparameter settings described in Table~\ref{tab:cura_vision_config}, \ours{} achieves a higher accuracy of 91.2\% with only 1.41M parameters. In contrast, ResMLP-S12~\cite{touvron2022resmlp} requires 10.9$\times$ more parameters (15.4M vs. 1.41M) while achieving a significantly lower accuracy of 81.66\%—a gap of 9.54 percentage points. EfficientNet-B0~\cite{tan2019efficientnet} achieves slightly higher accuracy (92.34\%) but uses 2.9$\times$ more parameters (4.02M), offering a marginal 1.14 percentage point improvement at much higher computational cost. Similarly, MobileNet-V2~\cite{sandler2018mobilenetv2} uses 1.6$\times$ more parameters (2.24M) but still falls short in accuracy, achieving only 89.33\%. These comparisons clearly demonstrate that \ours{} provides the best trade-off between accuracy and model complexity, which outperfors or matches larger models with a significantly more compact architecture.



\subsection{Assessment of Complex Pattern Forecast}
\begin{figure*}[t!]
    \centering
    \subfigure[Comparison of forecasting on S\&P 500 dataset.]{
        \includegraphics[width=7.2cm]{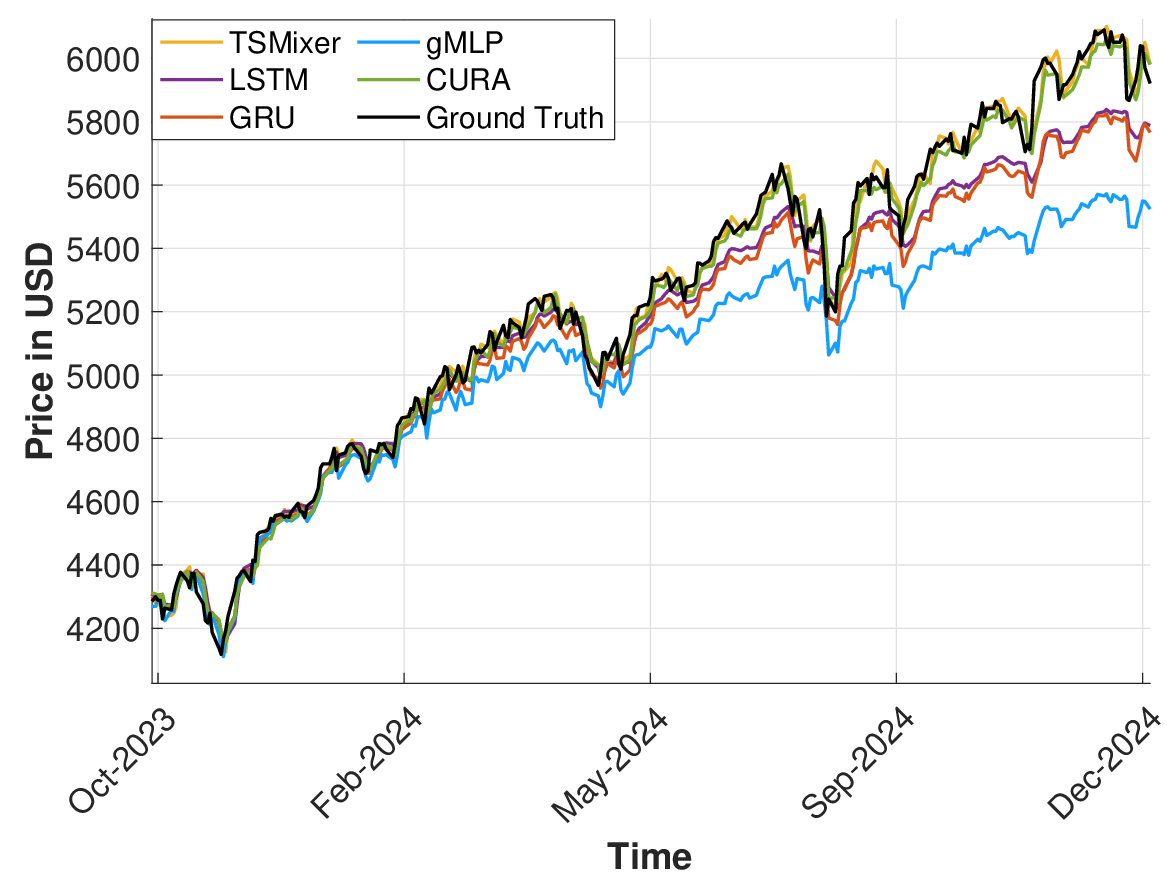}
        \label{fig:comp_sp_a}
    }
    \hspace{2cm}
    \subfigure[Comparison of forecasting on S\&P and ETTm1 dataset across different models.]{
        \includegraphics[width=7.5cm]{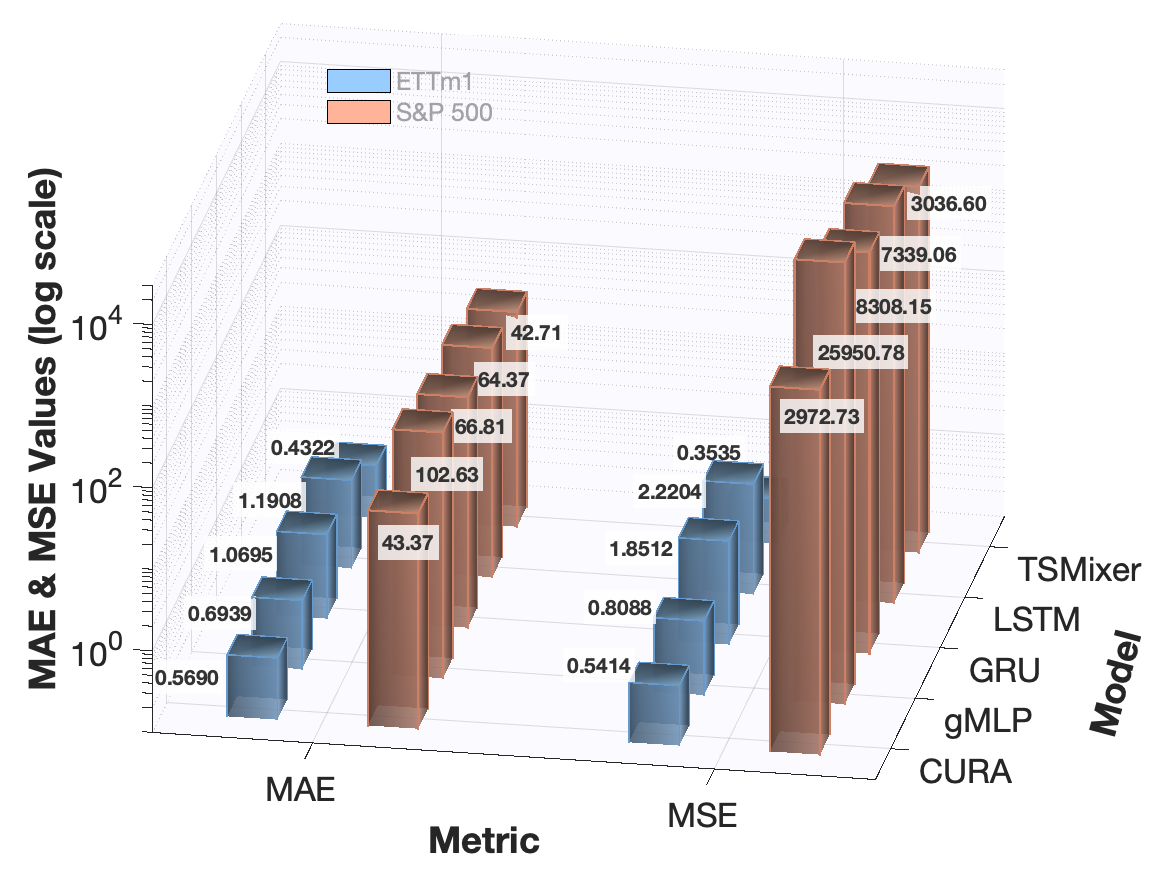}
        \label{fig:comp_sp_b}
    }
    \caption{Model performance comparison on two datasets: (a) S\&P 500 and (b) ETTm1.}
    \label{fig:comp_sp}
\end{figure*}

To assess the prediction accuracy of \ours{} in time series data, we compare it with various baseline models, namely,  
gMLP, GRU, LSTM, and TSMixer.
We initially evaluated our approach using S\&P 500 time series data.
The forecasting performance of each model is illustrated in Figure \ref{fig:comp_sp_a}. The black line represents the ground truth, showing the actual S\&P 500 market prices over the period from October 2023 to December 2024. Note that we display only a subset of the data to better visualize and compare the overall forecasting trends of each competing model.
For each model, we utilized historical closing prices from the previous 20 business days to predict the closing price of the following trading day. The prediction target is the daily closing price value, which corresponds to the "close" column in the dataset\footnote{Before training, these prices are standardized using z-score normalization, which centers the data around zero and scales them according to their standard deviation. This preprocessing step improves the efficiency of model training and the numerical stability. Once predictions are generated, they are converted back to the original price scale using the inverse transformation, which allows the results to be interpreted directly as actual market prices.}. Each model employs a sliding window that moves forward by one day for each new prediction sample. The dataset is split chronologically, with the first 80\% used for training to capture diverse historical market patterns and the remaining 20\% reserved for testing on genuinely unseen future data.

It can be observed that the ground truth (black line) is closely tracked by the green line (representing \ours{}) and the yellow line (representing TSMixer). Although TSMixer achieves the closest alignment with the ground truth, this performance comes at the cost of significantly higher parameter overhead. It is also notable that LSTM and GRU perform well for short-term forecasting; however, as the prediction window extends, they increasingly deviate from the ground truth. This occurs because these models rely on patterns from the past 20 days, and once prediction errors emerge, they compound and propagate, leading to progressively larger deviations. The same applies to gMLP that has the weakest performance due to its reliance on static permutation matrices for feature mixing, which cannot adequately model the dynamic temporal correlations required for extended forecasting horizons in financial time series data.

In addition to S\&P market prediction, we used the ETTm1 dataset to compare \ours{} with the aforementioned baseline models, as it contains data typically collected by edge devices. This dataset serves as a more suitable benchmark because it reflects the resource-constrained environments where our method is specifically designed to operate.
 We implemented a multivariate time series forecasting framework that employs a sliding-window approach to effectively capture the complex temporal dependencies inherent in the electricity transformer measurements.
 Each input sample consists of 96 consecutive time steps that include all 7 features available in the dataset. This 96-step window provides adequate historical context to capture both short-term fluctuations and longer-term operational trends typical of electrical transformers. The multivariate approach allows the model to exploit cross-correlations between different operational parameters, thereby enhancing prediction accuracy.
We designed the prediction task to forecast the next 24 time steps of the 
oil temperature 
variable, 
as it is a key indicator of transformer health and operational efficiency.
%
This time-frame provides adequate lead time for preventive maintenance decisions while maintaining reasonable accuracy throughout the forecast period.
%

\begin{table}[!t]
    \centering
    \caption{F1 Score and Parameters Comparison on CIFAR-10 Without Heavy Pretraining}
    \begin{tabular}{p{3cm} p{2.2cm} p{2.1cm}}
        \toprule
        \textbf{Model} & \textbf{No. of Params} & \textbf{F1 (\%)} \\
        \midrule\midrule
        CURA & \textbf{1.41M} & \textbf{91.2} \\
        ResMLP-S12 ~\cite{touvron2022resmlp} & 15.4M & 81.66 \\ 
        MobileNet-V2\cite{sandler2018mobilenetv2} & 2.24M & 89.33 \\
        EfficientNet-B0\cite{tan2019efficientnet} & 4.02M & 92.34 \\
        ViT-tiny\cite{dosovitskiy2020image} & 5.53M & 82.3\\
        \bottomrule
    \end{tabular}
    \label{tab:cifar10-cura-reformat}
\end{table}

Figure~\ref{fig:comp_sp_b} depicts the mean absolute error (MAE) and mean squared error (MSE) of various models, shown on a logarithmic scale, for the S\&P and ETTm1 datasets. 
On the ETTm1 dataset, it is observed that \ours{} provides competitive performance, outperforming all models except TSMixer. 
In the case of the S\&P 500 dataset, \ours{} shows slightly better performance than TSMixer in terms of MSE, while maintaining a marginally higher MAE. 
Despite the slight difference in performance, \ours{} operates with substantially fewer parameters compared to all the other models.
To further elaborate on these results, 
we summarize the MAE and MSE values alongside the parameter counts for all models considered in Table~\ref{tab:time_series_comparison}.

\vspace{0.25cm}
\noindent\textbf{Performance on ETTm1 dataset:}  
As shown in Table~\ref{tab:time_series_comparison}, TSMixer achieves the best predictive performance, attaining the lowest MAE and MSE among all models, albeit at the cost of employing 75 times more parameters than \ours{} (55,090 vs. 731). In contrast, \ours{} delivers the second-best performance, achieving comparably low MAE and MSE with only 731 parameters. Other models such as gMLP, GRU, and LSTM, while requiring 15 to 25 times more parameters than \ours{}, exhibit substantially higher errors, with MAE values between 22\% and 109\% greater than that of \ours{}. 
These findings demonstrate that, for the ETTm1 forecasting
task, increased model complexity does not necessarily translate
to better predictive performance. This indicates that compact
architectures like \ours{} can be highly competitive.


\vspace{0.25cm}

\noindent\textbf{Performance on S\&P 500 Dataset:}  
\ours{} achieves competitive performance while maintaining the lowest parameter count. Compared to TSMixer, which uses 4.8 times more parameters, \ours{} delivers almost identical accuracy, with only a 1.5\% higher MAE. In contrast, recurrent architectures such as GRU and LSTM perform substantially worse, showing 54\% and 48\% higher MAE, respectively, while requiring around 6 times more parameters. Similarly, gMLP generalizes poorly to these financial data, with 137\% higher MAE despite using 1.4 times more parameters than \ours{}.

\begin{table*}[!ht]
    \centering
    \caption{Performance Comparison on Time Series Forecasting Tasks}
    \begin{tabular}{p{2cm} p{2.5cm} p{2cm} p{2cm} p{2.5cm} p{2cm} p{2cm}}
        \toprule
        \multirow{2}{*}{\textbf{Model}} & \multicolumn{3}{c}{\textbf{ETTm1}} & \multicolumn{3}{c}{\textbf{S\&P 500}} \\
        \cmidrule(lr){2-4} \cmidrule(lr){5-7}
        & \textbf{Params} & \textbf{MAE} & \textbf{MSE} & \textbf{Params} & \textbf{MAE} & \textbf{MSE} \\
        \midrule\midrule
            \textbf{\ours{}} & 731 & 0.5690 & 0.5414 & 746 & 43.37 & 2,972.73 \\
        gMLP & 11,423 & 0.6939 & 0.8088 & 1,067 & 102.63 & 25,950.78 \\
        GRU & 14,081 & 1.0695 & 1.8512 & 4,479 & 66.81 & 8,308.15 \\
        LSTM & 18,753 & 1.1908 & 2.2204 & 4,513 & 64.37 & 7,339.06 \\
        TSMixer & 55,090 & 0.4322 & 0.3535 & 3,585 & 42.71 & 3,036.60 \\
        \bottomrule
    \end{tabular}
    \label{tab:time_series_comparison}
\end{table*}

\section{Discussion}
\ours{} is a novel architecture optimized for AI applications in resource-constrained environments. This design philosophy explicitly targets deployment on platforms such as smartphones, IoT devices, and wearables, where computational resources, memory, and power consumption are limited.

\vspace{0.3cm}
\noindent\textbf{CURA as a Computational Core.} 
The conventional deep learning modules are structured as blocks, which typically require substantial computational resources, memory bandwidth, and parameter storage due to their deep layer stacking, high-dimensional feature representations, and complex interconnections. These architectures often scale linearly or exponentially with task complexity, making them unsuitable for deployment in resource-constrained environments.
In contrast, \ours{} is designed as a lightweight computational core, which exhibits the following characteristics: (1) It is effective even without pretraining or deep stacking in basic tasks, meaning that the architecture can achieve competitive performance through end-to-end training from scratch. This eliminates the computational overhead and storage requirements associated with large pretrained models.
(2) It supports both parallel and sequential connections due to a consistent architectural interface that enables its seamless integration into diverse network topologies and allowing for dynamic reconfiguration based on computational constraints and task requirements.
(3) It allows flexible scalability across different model sizes and complexities because of its modular core structure and parameter-efficient design which enables its deployment ranging from ultra-lightweight edge applications to more complex multi-domain tasks without architectural redesign.
(4) It achieves strong results while maintaining minimal model size which results in a competitive accuracy with significantly reduced memory footprint and computational complexity compared to conventional deep learning approaches.

    
    
    
    


\vspace{0.3cm}
\noindent\textbf{Architectural Scalability and Modularity.} 
One representative application of \ours{} involves placing multiple \ours{} cores in parallel, merging their outputs into a single tensor, and feeding it to the next stage. This approach allows scalable performance without increasing the parameter count, which makes \ours{} an ideal computational core for efficient and modular deep learning architectures. This design paradigm enables practitioners to adapt computational capacity dynamically based on available resources while maintaining architectural consistency across different deployment scenarios. Nevertheless, at this point we suspects that 
the approach might introduces complexity in gradient synchronization during training, hyperparameter optimization across multiple cores, and maintaining energy efficiency on resource-constrained devices. In future work, we will be assessing these factors and would focus on developing adaptive merging techniques, dynamic load balancing algorithms, and hardware-aware scheduling mechanisms to realize the full potential of parallel \ours{} deployments while preserving the architecture's efficiency advantages.

\vspace{0.3cm}
\noindent\textbf{Empirical Validation Across Domains.} 
Our empirical evaluations of \ours{} across a wide range of tasks, including structured regression, time-series forecasting, natural language processing, and computer vision, demonstrate its exceptional parameter efficiency and competitive performance. These results validate \ours{}'s versatility as a unified computational primitive capable of adapting to heterogeneous input modalities and task requirements without architectural modifications, a crucial advantage for resource-constrained deployment scenarios. At this point we believe that \ours{} achieves competitive results on standard benchmarks. However, it may struggle with tasks requiring extensive world knowledge or sophisticated reasoning capabilities that large language models excel at through massive parameter counts and pretraining data. Assessing such capabilities remains our future work. 

\vspace{1em}
\section{Conclusion} 
This paper proposes \ours{}, an 
lightweight
architecture for resource-constrained devices that delivers three key advantages over conventional models. First, it achieves exceptional parameter efficiency, consuming significantly fewer parameters than existing models while maintaining equivalent task performance. Second, it demonstrates high generalizability, extending beyond traditional regression tasks to handle diverse NLP and computer vision classification problems. Third, it can capture intricate patterns with high accuracy while maintaining minimal model complexity.
We comprehensively evaluated these three objectives across 10 different datasets: five time-series sequential datasets and five cross-domain classification datasets. To assess model compactness, we evaluated \ours{} against popular baselines including gMLP, GRU, LSTM, and TSMaxer on time-series datasets. \ours{} achieved F1-scores up to 95\% compared to baseline models while using $2{,}500\times$ fewer parameters.
For generalizability assessment, we evaluated \ours{} on diverse NLP and vision tasks, including 4-way multiple-choice reasoning, multi-class classification, binary sentiment classification, and contextual binary QA. Compared to state-of-the-art baselines, \ours{} achieved up to 90\% accuracy using $21\times$ fewer parameters.
Finally, we assessed pattern forecasting capabilities on two specialized datasets: an on-device power system parameter collection dataset and a generalized marketing forecasting dataset. \ours{} demonstrated the lowest mean squared error (2.1× lower) and mean absolute error (1.6× lower), while using approximately 48× fewer parameters and maintaining comparable forecasting accuracy to competing methods.
These results validate \ours{} as a versatile, parameter-efficient solution for edge AI applications across multiple domains. We have open-sourced our architecture code and supplementary experiments in our GitHub 
%
repository to facilitate reproducibility and future research
(https://github.com/SION001122/CURA).

\bibliographystyle{IEEEtran}
\bibliography{Bibliography}

\begin{thebibliography}{10}
\providecommand{\url}[1]{#1}
\csname url@samestyle\endcsname
\providecommand{\newblock}{\relax}
\providecommand{\bibinfo}[2]{#2}
\providecommand{\BIBentrySTDinterwordspacing}{\spaceskip=0pt\relax}
\providecommand{\BIBentryALTinterwordstretchfactor}{4}
\providecommand{\BIBentryALTinterwordspacing}{\spaceskip=\fontdimen2\font plus
\BIBentryALTinterwordstretchfactor\fontdimen3\font minus \fontdimen4\font\relax}
\providecommand{\BIBforeignlanguage}[2]{{%
\expandafter\ifx\csname l@#1\endcsname\relax
\typeout{** WARNING: IEEEtran.bst: No hyphenation pattern has been}%
\typeout{** loaded for the language `#1'. Using the pattern for}%
\typeout{** the default language instead.}%
\else
\language=\csname l@#1\endcsname
\fi
#2}}
\providecommand{\BIBdecl}{\relax}
\BIBdecl

\bibitem{silvano2025survey}
C.~Silvano, D.~Ielmini, F.~Ferrandi, L.~Fiorin, S.~Curzel, L.~Benini, F.~Conti, A.~Garofalo, C.~Zambelli, E.~Calore \emph{et~al.}, ``A survey on deep learning hardware accelerators for heterogeneous hpc platforms,'' \emph{ACM Computing Surveys}, vol.~57, no.~11, pp. 1--39, 2025.

\bibitem{liang2020ai}
Q.~Liang, P.~Shenoy, and D.~Irwin, ``Ai on the edge: Characterizing ai-based iot applications using specialized edge architectures,'' in \emph{2020 IEEE International symposium on workload characterization (IISWC)}.\hskip 1em plus 0.5em minus 0.4em\relax IEEE, 2020, pp. 145--156.

\bibitem{thota2024optimizing}
R.~C. Thota, ``Optimizing edge computing and ai for low-latency cloud workloads,'' \emph{International Journal of Science and Research Archive}, vol.~13, no.~1, pp. 3484--3500, 2024.

\bibitem{dhar2021survey}
S.~Dhar, J.~Guo, J.~Liu, S.~Tripathi, U.~Kurup, and M.~Shah, ``A survey of on-device machine learning: An algorithms and learning theory perspective,'' \emph{ACM Transactions on Internet of Things}, vol.~2, no.~3, pp. 1--49, 2021.

\bibitem{tekin2024review}
N.~Tekin, A.~Aris, A.~Acar, S.~Uluagac, and V.~C. Gungor, ``A review of on-device machine learning for iot: An energy perspective,'' \emph{Ad Hoc Networks}, vol. 153, p. 103348, 2024.

\bibitem{shi2020communication}
Y.~Shi, K.~Yang, T.~Jiang, J.~Zhang, and K.~B. Letaief, ``Communication-efficient edge ai: Algorithms and systems,'' \emph{IEEE Communications Surveys \& Tutorials}, vol.~22, no.~4, pp. 2167--2191, 2020.

\bibitem{sandler2018mobilenetv2}
M.~Sandler, A.~Howard, M.~Zhu, A.~Zhmoginov, and L.-C. Chen, ``Mobilenetv2: Inverted residuals and linear bottlenecks,'' in \emph{Proceedings of the IEEE conference on computer vision and pattern recognition}, 2018, pp. 4510--4520.

\bibitem{jiao-etal-2020-tinybert}
X.~Jiao, Y.~Yin, L.~Shang, X.~Jiang, X.~Chen, L.~Li, F.~Wang, and Q.~Liu, ``{T}iny{BERT}: Distilling {BERT} for natural language understanding,'' in \emph{Findings of the Association for Computational Linguistics: EMNLP 2020}, T.~Cohn, Y.~He, and Y.~Liu, Eds.\hskip 1em plus 0.5em minus 0.4em\relax Online: Association for Computational Linguistics, Nov. 2020, pp. 4163--4174.

\bibitem{NIPS2017_3f5ee243}
A.~Vaswani, N.~Shazeer, N.~Parmar, J.~Uszkoreit, L.~Jones, A.~N. Gomez, L.~u. Kaiser, and I.~Polosukhin, ``Attention is all you need,'' in \emph{Advances in Neural Information Processing Systems}, I.~Guyon, U.~V. Luxburg, S.~Bengio, H.~Wallach, R.~Fergus, S.~Vishwanathan, and R.~Garnett, Eds., vol.~30.\hskip 1em plus 0.5em minus 0.4em\relax Curran Associates, Inc., 2017.

\bibitem{NEURIPS2021_4cc05b35}
H.~Liu, Z.~Dai, D.~So, and Q.~V. Le, ``Pay attention to mlps,'' in \emph{Advances in Neural Information Processing Systems}, M.~Ranzato, A.~Beygelzimer, Y.~Dauphin, P.~Liang, and J.~W. Vaughan, Eds., vol.~34.\hskip 1em plus 0.5em minus 0.4em\relax Curran Associates, Inc., 2021, pp. 9204--9215.

\bibitem{touvron2022resmlp}
H.~Touvron, P.~Bojanowski, M.~Caron, M.~Cord, A.~El-Nouby, E.~Grave, G.~Izacard, A.~Joulin, G.~Synnaeve, J.~Verbeek \emph{et~al.}, ``Resmlp: Feedforward networks for image classification with data-efficient training,'' \emph{IEEE transactions on pattern analysis and machine intelligence}, vol.~45, no.~4, pp. 5314--5321, 2022.

\bibitem{chen2023tsmixer}
S.-A. Chen, C.-L. Li, N.~Yoder, S.~O. Arik, and T.~Pfister, ``Tsmixer: An all-mlp architecture for time series forecasting,'' \emph{arXiv preprint arXiv:2303.06053}, 2023.

\bibitem{ekambaram2023tsmixer}
V.~Ekambaram, A.~Jati, N.~Nguyen, P.~Sinthong, and J.~Kalagnanam, ``Tsmixer: Lightweight mlp-mixer model for multivariate time series forecasting,'' in \emph{Proceedings of the 29th ACM SIGKDD conference on knowledge discovery and data mining}, 2023, pp. 459--469.

\bibitem{hochreiter1997long}
S.~Hochreiter and J.~Schmidhuber, ``Long short-term memory,'' \emph{Neural computation}, vol.~9, no.~8, pp. 1735--1780, 1997.

\bibitem{karpathy2015visualizing}
A.~Karpathy, J.~Johnson, and L.~Fei-Fei, ``Visualizing and understanding recurrent networks,'' \emph{arXiv preprint arXiv:1506.02078}, 2015.

\bibitem{pmlr-v37-jozefowicz15}
R.~Jozefowicz, W.~Zaremba, and I.~Sutskever, ``An empirical exploration of recurrent network architectures,'' in \emph{Proceedings of the 32nd International Conference on Machine Learning}, ser. Proceedings of Machine Learning Research, F.~Bach and D.~Blei, Eds., vol.~37.\hskip 1em plus 0.5em minus 0.4em\relax Lille, France: PMLR, 07--09 Jul 2015, pp. 2342--2350.

\bibitem{cho-etal-2014-learning}
K.~Cho, B.~van Merri{\"e}nboer, C.~Gulcehre, D.~Bahdanau, F.~Bougares, H.~Schwenk, and Y.~Bengio, ``Learning phrase representations using {RNN} encoder{--}decoder for statistical machine translation,'' in \emph{Proceedings of the 2014 Conference on Empirical Methods in Natural Language Processing ({EMNLP})}, A.~Moschitti, B.~Pang, and W.~Daelemans, Eds.\hskip 1em plus 0.5em minus 0.4em\relax Doha, Qatar: Association for Computational Linguistics, Oct. 2014, pp. 1724--1734.

\bibitem{tan2019efficientnet}
M.~Tan and Q.~Le, ``Efficientnet: Rethinking model scaling for convolutional neural networks,'' in \emph{International conference on machine learning}.\hskip 1em plus 0.5em minus 0.4em\relax PMLR, 2019, pp. 6105--6114.

\bibitem{howard2019searching}
A.~Howard, M.~Sandler, G.~Chu, L.-C. Chen, B.~Chen, M.~Tan, W.~Wang, Y.~Zhu, R.~Pang, V.~Vasudevan \emph{et~al.}, ``Searching for mobilenetv3,'' in \emph{Proceedings of the IEEE/CVF international conference on computer vision}, 2019, pp. 1314--1324.

\bibitem{warden2019tinyml}
P.~Warden and D.~Situnayake, \emph{Tinyml: Machine learning with tensorflow lite on arduino and ultra-low-power microcontrollers}.\hskip 1em plus 0.5em minus 0.4em\relax O'Reilly Media, 2019.

\bibitem{capogrosso2024machine}
L.~Capogrosso, F.~Cunico, D.~S. Cheng, F.~Fummi, and M.~Cristani, ``A machine learning-oriented survey on tiny machine learning,'' \emph{IEEE Access}, vol.~12, pp. 23\,406--23\,426, 2024.

\bibitem{nosouhian2021review}
S.~Nosouhian, F.~Nosouhian, and A.~K. Khoshouei, ``A review of recurrent neural network architecture for sequence learning: Comparison between lstm and gru,'' 2021.

\bibitem{yamak2019comparison}
P.~T. Yamak, L.~Yujian, and P.~K. Gadosey, ``A comparison between arima, lstm, and gru for time series forecasting,'' in \emph{Proceedings of the 2019 2nd international conference on algorithms, computing and artificial intelligence}, 2019, pp. 49--55.

\bibitem{anguita2013public}
D.~Anguita, A.~Ghio, L.~Oneto, X.~Parra, J.~L. Reyes-Ortiz \emph{et~al.}, ``A public domain dataset for human activity recognition using smartphones.'' in \emph{Esann}, vol.~3, no.~1, 2013, pp. 3--4.

\bibitem{Zhou_Zhang_Peng_Zhang_Li_Xiong_Zhang_2021}
H.~Zhou, S.~Zhang, J.~Peng, S.~Zhang, J.~Li, H.~Xiong, and W.~Zhang, ``Informer: Beyond efficient transformer for long sequence time-series forecasting,'' \emph{Proceedings of the AAAI Conference on Artificial Intelligence}, vol.~35, no.~12, pp. 11\,106--11\,115, May 2021.

\bibitem{saleh10fallalld}
M.~Saleh and R.~L.~B. Jeannes, ``Fallalld: a comprehensive dataset of human falls and activities of daily living, 2020,'' \emph{DOI: https://doi. org/10.21227/bnya-mn34}.

\bibitem{houseprices}
``House prices - advanced regression techniques,'' \url{https://www.kaggle.com/competitions/house-prices-advanced-regression-techniques}, 2016, accessed: June 11, 2025.

\bibitem{aroussi2025yfinance}
R.~Aroussi, ``yfinance: Download market data from yahoo! finance’s api (version 0.2. 54)[software]. pypi,'' 2025.

\bibitem{krizhevsky2009learning}
A.~Krizhevsky, G.~Hinton \emph{et~al.}, ``Learning multiple layers of features from tiny images,'' 2009.

\bibitem{NIPS2015_250cf8b5}
X.~Zhang, J.~Zhao, and Y.~LeCun, ``Character-level convolutional networks for text classification,'' in \emph{Advances in Neural Information Processing Systems}, C.~Cortes, N.~Lawrence, D.~Lee, M.~Sugiyama, and R.~Garnett, Eds., vol.~28.\hskip 1em plus 0.5em minus 0.4em\relax Curran Associates, Inc., 2015.

\bibitem{clark-etal-2019-boolq}
C.~Clark, K.~Lee, M.-W. Chang, T.~Kwiatkowski, M.~Collins, and K.~Toutanova, ``{B}ool{Q}: Exploring the surprising difficulty of natural yes/no questions,'' in \emph{Proceedings of the 2019 Conference of the North {A}merican Chapter of the Association for Computational Linguistics: Human Language Technologies, Volume 1 (Long and Short Papers)}, J.~Burstein, C.~Doran, and T.~Solorio, Eds.\hskip 1em plus 0.5em minus 0.4em\relax Minneapolis, Minnesota: Association for Computational Linguistics, Jun. 2019, pp. 2924--2936.

\bibitem{zellers-etal-2019-hellaswag}
R.~Zellers, A.~Holtzman, Y.~Bisk, A.~Farhadi, and Y.~Choi, ``{H}ella{S}wag: Can a machine really finish your sentence?'' in \emph{Proceedings of the 57th Annual Meeting of the Association for Computational Linguistics}, A.~Korhonen, D.~Traum, and L.~M{\`a}rquez, Eds.\hskip 1em plus 0.5em minus 0.4em\relax Florence, Italy: Association for Computational Linguistics, Jul. 2019, pp. 4791--4800.

\bibitem{lewis2019bart}
M.~Lewis, Y.~Liu, N.~Goyal, M.~Ghazvininejad, A.~Mohamed, O.~Levy, V.~Stoyanov, and L.~Zettlemoyer, ``Bart: Denoising sequence-to-sequence pre-training for natural language generation, translation, and comprehension,'' \emph{arXiv preprint arXiv:1910.13461}, 2019.

\bibitem{dosovitskiy2020image}
A.~Dosovitskiy, L.~Beyer, A.~Kolesnikov, D.~Weissenborn, X.~Zhai, T.~Unterthiner, M.~Dehghani, M.~Minderer, G.~Heigold, S.~Gelly \emph{et~al.}, ``An image is worth 16x16 words: Transformers for image recognition at scale,'' \emph{arXiv preprint arXiv:2010.11929}, 2020.

\end{thebibliography}

\end{document}